\documentclass[journal,article,submit,pdftex,moreauthors]{Definitions/mdpi}
\firstpage{1} 
\pubvolume{1}
\issuenum{1}
\articlenumber{0}
\pubyear{2023}
\copyrightyear{2023}
\datereceived{ } 
\daterevised{ } 
\dateaccepted{ } 
\datepublished{ } 
\hreflink{https://doi.org/} 
\usepackage[misc,geometry]{ifsym}
\usepackage{xcolor}
\usepackage{CJKutf8}

\Title{TransGPT: Multi-modal Generative Pre-trained Transformer for Transportation}

\Author{Peng Wang $^{1}$, Xiang Wei $^{1}$, Fangxu Hu $^{1}$ and Wenjuan Han$^{1\,\textrm{\Letter}}$}

\address{%
$^{1}$ \quad Beijing Jiaotong University, Beijing, China\\
\Letter\quad Correspondence: Wenjuan Han; wjhan@bjtu.edu.cn}

\abstract{Natural language processing (NLP) is a key component of intelligent transportation systems (ITS), but it faces many challenges in the transportation domain, such as domain-specific knowledge and data, and multi-modal inputs and outputs. This paper presents TransGPT, a novel (multi-modal) large language model for the transportation domain, which consists of two independent variants: TransGPT-SM for single-modal data and TransGPT-MM for multi-modal data. TransGPT-SM is finetuned on a single-modal Transportation dataset (STD) that contains textual data from various sources in the transportation domain. TransGPT-MM is finetuned on a multi-modal Transportation dataset (MTD) that we manually collected from three areas of the transportation domain: driving tests, traffic signs, and landmarks. We evaluate TransGPT on several benchmark datasets for different tasks in the transportation domain, and show that it outperforms baseline models on most tasks. We also showcase the potential applications of TransGPT for traffic analysis and modeling, such as generating synthetic traffic scenarios, explaining traffic phenomena, answering traffic-related questions, providing traffic recommendations, and generating traffic reports. This work advances the state-of-the-art of NLP in the transportation domain and provides a useful tool for ITS researchers and practitioners.}


\newcommand{\yourname}{ }
\newcommand{\youremail}{ }
\usepackage{fancyhdr}
\begin{document}
\thispagestyle{fancy}



\section{Introduction}
Transportation is a vital aspect of modern society, affecting the economy, the environment, and the quality of life of billions of people. Meanwhile, transportation also poses many challenges, such as congestion, pollution, safety, and efficiency. To address these challenges, there is a need for intelligent transportation systems (ITS) that can analyze, predict, and optimize transportation-related states and behaviors.
AI enhancing transportation systems is not a trivial task, as it requires a deep understanding of the domain-specific concepts, rules, regulations, policies, and best practices, as well as the ability to handle the complexity and diversity of real-world transportation scenarios.

In recent years, large language model-based Artificial Intelligence systems have augmented humans in certain roles~\cite{vaswani2017attention,sanh2021multitask,ouyang2022training, zhang2022opt}, and soon this trend has expanded to the vast majority of numerous fields, for example medicine~\cite{nori2023capabilities}, law~\cite{choi2023chatgpt} and finance~\cite{wu2023bloomberggpt}. Large language models (LLMs) are deep neural networks that can learn from massive amounts of text data. LLMs can capture the semantic and syntactic patterns of human language and generate coherent and fluent texts for various applications and domains~\cite{shao2023compositional,NEURIPS20201457c0d6}.

However, most LLMs are pre-trained on general corpora, such as Common Crawl and Wikipedia, which may not cover the specificities and nuances of the transportation domain. Therefore, applying LLMs directly to the transportation domain may result in inaccurate or irrelevant outputs. To overcome this limitation, there is a need for domain-specific LLMs that can leverage the domain knowledge and data to perform better in the transportation domain.

In this paper, we propose TransGPT\footnote{Website: \url{https://github.com/DUOMO/TransGPT}}, a novel (V)LLM for the transportation domain that consists of two variants: single-modal variant (TransGPT-SM), and multi-modal variant (TransGPT-MM).
TransGPT-SM is a finetuned version of ChatGLM2-6B~\cite{du2021glm}, which is a large language model for bilingual (Chinese and English) chat. TransGPT-SM is finetuned on a single-modal Transportation dataset (STD) that contains 12.5 million tokens of textual data collected from various sources in the transportation domain, such as books, reports, documents, websites, and corpora. TransGPT-SM can generate natural language outputs for various transportation-related tasks based on textual inputs.
TransGPT-MM is based on VisualGLM-6B\footnote{\url{https://github.com/THUDM/VisualGLM-6B}}, which is an open-source multi-modal LLM that supports images and texts. TransGPT-MM is finetuned on a multi-modal Transportation dataset (MTD) that we manually collected from various sources in three areas of the transportation domain: driving tests, traffic signs, typical landmark, etc. The MTD consists of aligned images and texts, where the texts are divided into questions and answers. For finetuning, image and question are used as input and answer as the golden output.
We assess TransGPT on curated benchmark datasets for various tasks in the transportation field. Our experiments demonstrate that TransGPT achieves superior performance compared to ChatGLM2-6B, VisualGLM-6B, and other baseline models across the majority of tasks.

The key contributions of this paper can be summarized as follows:
\begin{itemize}
    \item We introduce TransGPT, a novel (V)LLM for the transportation domain that consists of two variants: TransGPT-SM and TransGPT-MM.
    \item We present STD and MTD, two datasets that we manually collected from various sources in the transportation domain: STD for single-modal data and MTD for multi-modal data.
    \item We evaluate TransGPT on the transportation benchmark. Our experiments show that TransGPT outperforms ChatGLM2-6B, VisualGLM-6B, and other baseline models in most tasks.
\end{itemize}
 
The rest of this paper is organized as follows: Section \ref{sec:related} reviews the related work on non-transformer-based and transformer-based models for transportation systems. Section \ref{sec:method} deceits construction of TransGPT. It includes three subsections: data collection, base model selection, and training. The section of data collection introduces the data collection methods and the characteristics of the datasets, including single-modal finetuning dataset and multi-modal finetuning dataset. The section of base model selection introduces the model selection criteria and the model structure details. The section of training introduces the training process and methods, for both single-modal version (TransGPT-SM) and multi-modal version (TransGPT-MM). Section \ref{sec:exp} introduces the experimental settings and results, including single-modal and multi-modal experiments. Section Section \ref{sec:analysis} provides a comprehensive analysis of the key steps and prompts related to TransGPT, shedding light on its significance and potential impact. 
Through an in-depth examination of the case study, this section delves into the intricacies of TransGPT, exploring its strengths and weaknesses.
Section \ref{sec:conclusion} concludes this paper and suggests future directions. 
\section{Related Work}\label{sec:related}
\subsection{Non-transformer-based Models for Transportation Systems}
Transportation systems, or intelligent transportation systems, are crucial for supporting decision-making among traffic managers, normal users, and policymakers. 
Machine learning (ML) techniques typically function as the brain of Intelligent Transportation Systems (ITS), with their accuracy and reliability. Traditional ML methods such as Support Vector Machine (SVM), Bayesian Network (BN), and Kalman Filter (KF) were initially used, but have since been revolutionized by the introduction of various deep learning models. These deep learning models have consistently achieved groundbreaking levels of accuracy in numerous transportation applications. It is therefore common practice to use deep learning models as predictors in ITS to improve accuracy. We can categorize the applications in ITS that rely on accurate learning models into visual recognition tasks, traffic flow prediction (TFP)~\cite{xiaojian2009traffic,huang2013adaptive}, traffic speed prediction (TSP)~\cite{lemieux2015vehicle,jia2016traffic}, travel time prediction (TTP)~\cite{gang2015continuous,siripanpornchana2016travel,duan2016travel}, and Miscellaneous tasks~\cite{genders2016using,tripathi2019convolutional}. For example, several widely recognized deep learning models have been developed through research and practical experience in the field of transportation, network topology analysis~\cite{kwayu2021discovering,osei2014complex}, network volume analysis~\cite{willumsen1978estimation,lingras2000traffic}, and network performance analysis~\cite{fisk1991traffic,isradi2020performance,marisamynathan2016performance}. 
In addition to sophisticated models for data analysis, Chen et al.~\cite{chen2015survey} emphasize the importance of data visualization in understanding the behavior of traffic participants and identifying traffic patterns. Existing data visualization models~\cite{andrienko2008basic,andrienko2012visual} mainly present location-based, activity-based, and device-based traffic data, with a focus on various properties including temporal ~\cite{guo2011tripvista} and spatial~\cite{liu2013vait,xie2008kernel,andrienko2009interactive} information.

\subsection{Transformer-based Models for Transportation Systems}
Although task-specific models created for single-round inputs and outputs performs well, the existence of LLMs along with the potential to incorporate numerous models, offers a strong basis for implementing transportation LLMs to combine advantages of individual task-specific models, tackle intricate traffic-related problems and facilitate decision-making.
To handle numerical sub-tasks, it is necessary to create transportation LLMs. Previous researchers have dedicated their efforts to exploring the fundamental structure of LLMs, specifically Transformers. For example, ~\cite{xu2020spatial} propose a paradigm of Spatial-Temporal Transformer Networks that leverages dynamical directed spatial and temporal dependencies to improve the accuracy of long-term traffic forecasting.  
Informer~\cite{2021Informer} suggests using a self-attention mechanism as an alternative to the standard self-attention in order to improve the predictive ability of the standard Transformer for forecasting problems.
Autoformer~\cite{2021AutoFormer} introduces an efficient Auto-Correlation mechanism.
DLinear~\cite{2023DLinear} takes a different approach by reimagining Transformer-based techniques and proposing a simple linear model based on decomposition.
Lastly, ~\cite{deshmukh2023swin} proposes a swin transformer-based vehicle detection framework. However, they focuses on limited sub-tasks.
Our work expands the capabilities of LLM in handling complex traffic-related tasks involving large numerical datasets. This includes, but is not limited to, data processing, data visualization, visual recognition tasks, traffic flow prediction, traffic speed prediction, travel time prediction, and Miscellaneous task.

\section{Construction of TransGPT}\label{sec:method}
\subsection{Data Collection}
We collected data from various sources in the transportation domain\footnote{Our data is in Chinese, and all examples in this manuscript are presented in both Chinese and English for ease of understanding.}.
Initially, we briefly introduce the resource of our collected data.
Since there is no existing large-scale dataset specifically for the transportation domain, we manually collected data from various sources, such as books, reports, documents, websites, and corpora. All collected data is authorized, and sensitive information is removed. We divided the data sources into five categories:

\begin{itemize}
    \item \textit{Traffic Engineering Documents}: This category encompasses documents related to traffic engineering specifications, analyses, and technical architectures. The data primarily originates from transportation technology projects and basic information of experts in the field, sourced from government departments, research institutions, universities, and enterprises. These data provide insights into current research trends and challenges in the field of transportation.
    \item \textit{Theses}: This category includes papers on various aspects such as transportation planning, traffic assessment, and traffic forecasting. The data is mainly sourced from academic databases, websites, and libraries both domestically and internationally. These documents contribute to our understanding of theoretical research and empirical analyses in the field of transportation.
    \item \textit{Examination Documents}: This category comprises documents related to exams on topics like driving licenses, traffic regulations, and vocational tests. The data is primarily sourced from transportation management departments, educational institutions, and training organizations. These documents provide content for knowledge education and skill training in the transportation domain.
    \item \textit{Report Documents}: This category includes documents such as development reports and think tank reports within the transportation industry. The data is primarily sourced from industry associations, consulting firms, and media organizations. These documents offer information for strategic guidance and decision-making recommendations in the transportation industry.
    \item \textit{Others}: We compiled data related to transportation statistical terms and transportation infrastructure construction knowledge. We also used the transportation section of the Fudan University Chinese Corpus\footnote{\url{https://github.com/hanwei2008/MavenTest}} to enrich our data.
\end{itemize}
This raw data collected by us contains 97.6 million textual tokens and over 3,000 images, building our transportation database. Then we base on the transportation database and construct the data for training (Section \ref{sec:data-training}) and evaluation (Section \ref{sec:data-evaluation}).
We call the single-modal transportation dataset STD, and the multi-modal transportation dataset MTD.

\subsubsection{Data Collection for Training}\label{sec:data-training}
\paragraph{\textbf{Single-modal finetuning dataset}}


To facilitate the rapid acquisition of knowledge within a new domain by finetuning large-scale language models, the provisioning of substantial instructional data for finetuning is imperative, allowing the model to generate corresponding outputs based on provided instructions and inputs. However, the acquisition and annotation of such data prove to be exceedingly time-consuming, expensive, and challenging to cover all conceivable domains and tasks. To address this challenge, we propose a methodology, which leverages unlabeled textual data to autonomously generate instructional data. These instructional data are in the form of question-answer pairs. The specific procedural steps are as follows:

\begin{figure*}[htp]
    \centering
    \includegraphics[width=\linewidth]{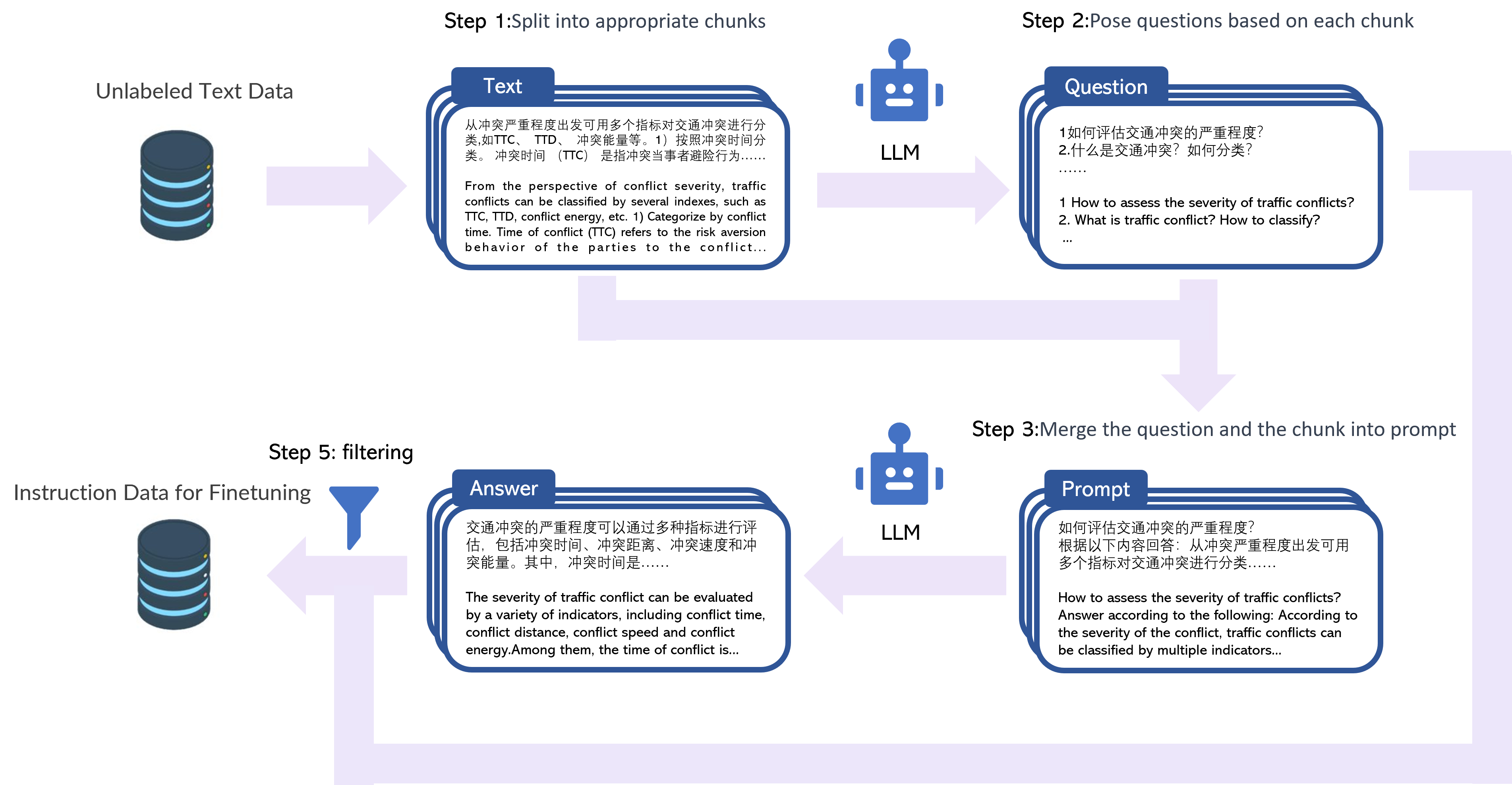}  
    \caption{Illustration of a flowchart showing how to automatically generate instruction data from unlabeled text data. The method consists of five steps. The flowchart uses arrows to indicate the sequence between steps, beginning in the upper left corner and ending in the lower left corner.}
    \label{fig:data-prompt-example}
\end{figure*}

\begin{itemize}
\item \textbf{Step 1}: Splitting the Original Document into Appropriate Chunks \\Initially, we segment the unsupervised textual data into suitable chunks, where each chunk encompasses a distinct theme. For instance, text division can be achieved using paragraph breaks, sentence boundaries, or section titles. \\
\item \textbf{Step 2}: Posing Questions Based on Each Chunk \\ Subsequently, leveraging a large language model, we generate one or multiple questions for each paragraph. These questions serve to test the model's comprehension and retention of the content within the respective paragraph. Furthermore, we utilize a question seed library to furnish examples, guiding the model in generating questions that align with our preferences. \\
\item \textbf{Step 3}:  Merging Questions and Chunks into Prompts\\
Following the question generation in Step 2, we appropriately merge the generated questions with the original document chunks, forming new prompts for the model. This fusion involves integrating the questions with the corresponding original document chunk, creating a coherent prompt with rich transporatation information. \\
\item \textbf{Step 4}: Answering Questions Based on Each Chunk
The subsequent step involves instructing the language model to generate responses to the questions posed, considering each paragraph's content. The generated answers are expected to be accurate, comprehensive, and reflective of the key information within the paragraph. To aid model identification, prefixes and suffixes may be employed to indicate the initiation and conclusion of questions and answers. \\
\item \textbf{Step 5}: Filtering and Merging Questions and Answers into Structured Text \\
In the final step, both questions and answers undergo filtering before being merged into structured text. Due to the absence of historical context during the question generation process, the model may sometimes repetitively pose similar questions or ask the same question about analogous text. We explicitly instruct in the prompt that if the model cannot answer based on the provided chunk or the query exceeds the model's comprehension scope, no output should be generated. In such cases, there might be instances of empty or invalid responses. Additionally, if the model deviates from instructions in its responses, those responses are also filtered out. After the filtering process, we proceed to integrate the question-answer pairs into structured text. This integrated text serves as instructional data for guiding subsequent model training endeavors.
\end{itemize}

In conclusion, following this series of steps, we ran Qwen-14B-chat\footnote{\url{https://huggingface.co/Qwen/Qwen-14B-Chat}} for 124 hours based on 8*A40 GPUs. After filtering, we ultimately obtained approximately 156,000 structured question-answer pairs for finetuning.



\begin{figure*}[htp]
    \centering
    \includegraphics[width=\linewidth]{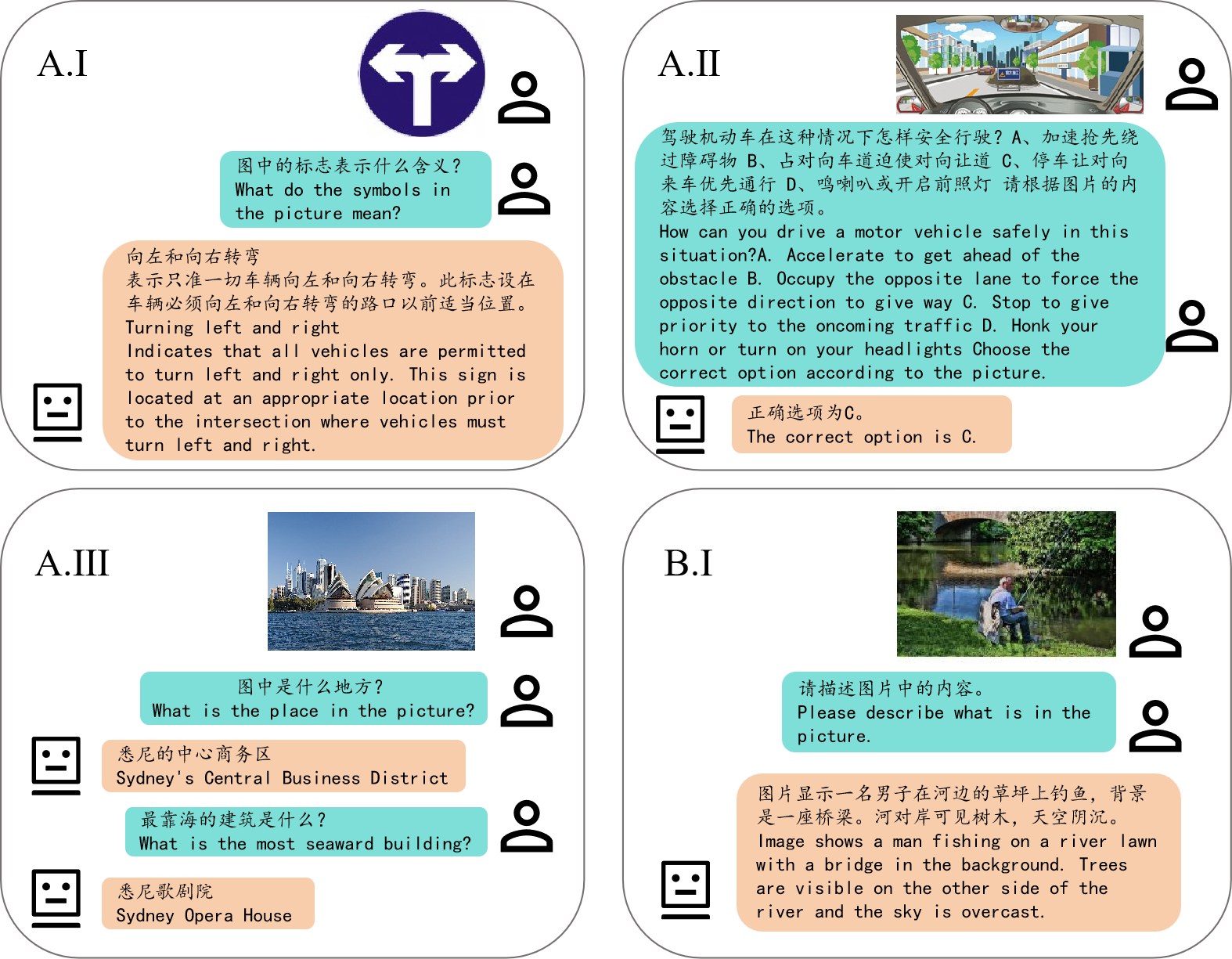}  
    \caption{Illustration of MTD and CCAC samples. A.I denotes \texttt{traffic signs}, A.II denotes \texttt{driving tests}, A.III denotes \texttt{landmarks}, B.I denotes \texttt{general image caption}.}
    \label{fig:mdt}
\end{figure*}

\paragraph{\textbf{Multi-modal finetuning dataset}}
To enable our model to handle both textual and visual inputs, we manually collected multi-modal resources in three areas of the transportation domain: driving tests, traffic signs, and landmarks. The resources provided are in various formats such as PPT, docx, and PDF. We began by extracting the images and texts from these formats. Next, we carefully aligned each image and text in the desired order to construct text-image samples.
We then devised a question for each sample. We call this multi-modal Transportation dataset MTD (Multi-modal Driving Test), containing 3183 samples for training and 122 samples for testing.
In addition, we take the second stage alignment dataset\footnote{\url{https://huggingface.co/datasets/Vision-CAIR/cc_sbu_align}} of MinGPT4~\cite{zhu2023minigpt} as the basis and use google translate\footnote{\url{https://translate.google.com/}} to convert it into a Chinese version, which we call CCAC containing 3439 samples. It is used in particular to maintain the generalized capabilities of TransGPT-MM.
In Figure \ref{fig:mdt}, we present some samples for MTD and CCAC.


\subsubsection{Data Collection for Evaluation}\label{sec:data-evaluation}
To evaluate the performance of our single-modal TransGPT-SM and multi-modal TransGPT-MM in the transportation domain, we use the following evaluation datasets called TransEval:

\begin{itemize}
    \item \textbf{Single-modal}: The evaluation data covers as many disciplines as possible in the transportation domain, such as traffic engineering, urban planning, traffic planning, public transportation, traffic management, traffic safety, etc. Since multiple-choice questions are a simple but good proxy for evaluating the potential of domain models for high-level abilities, we constructed all questions in multiple-choice format, with a total of 1237 questions.
    \item \textbf{Multi-modal}: We collected multi-modal resources manually in three areas of the transportation domain, namely driving test subjects, traffic signs, and landmarks. We aligned the images and texts sequentially to form samples, where the image serves as the input and the text serves as the corresponding output. Finally, we created a question-answer pair for each sample. We designed 122 samples for test.
\end{itemize}

\subsubsection{Data Analysis}
\begin{figure*}[htp]
    \centering
    \begin{minipage}{0.48\linewidth}
        \centering
        \includegraphics[width=\linewidth]{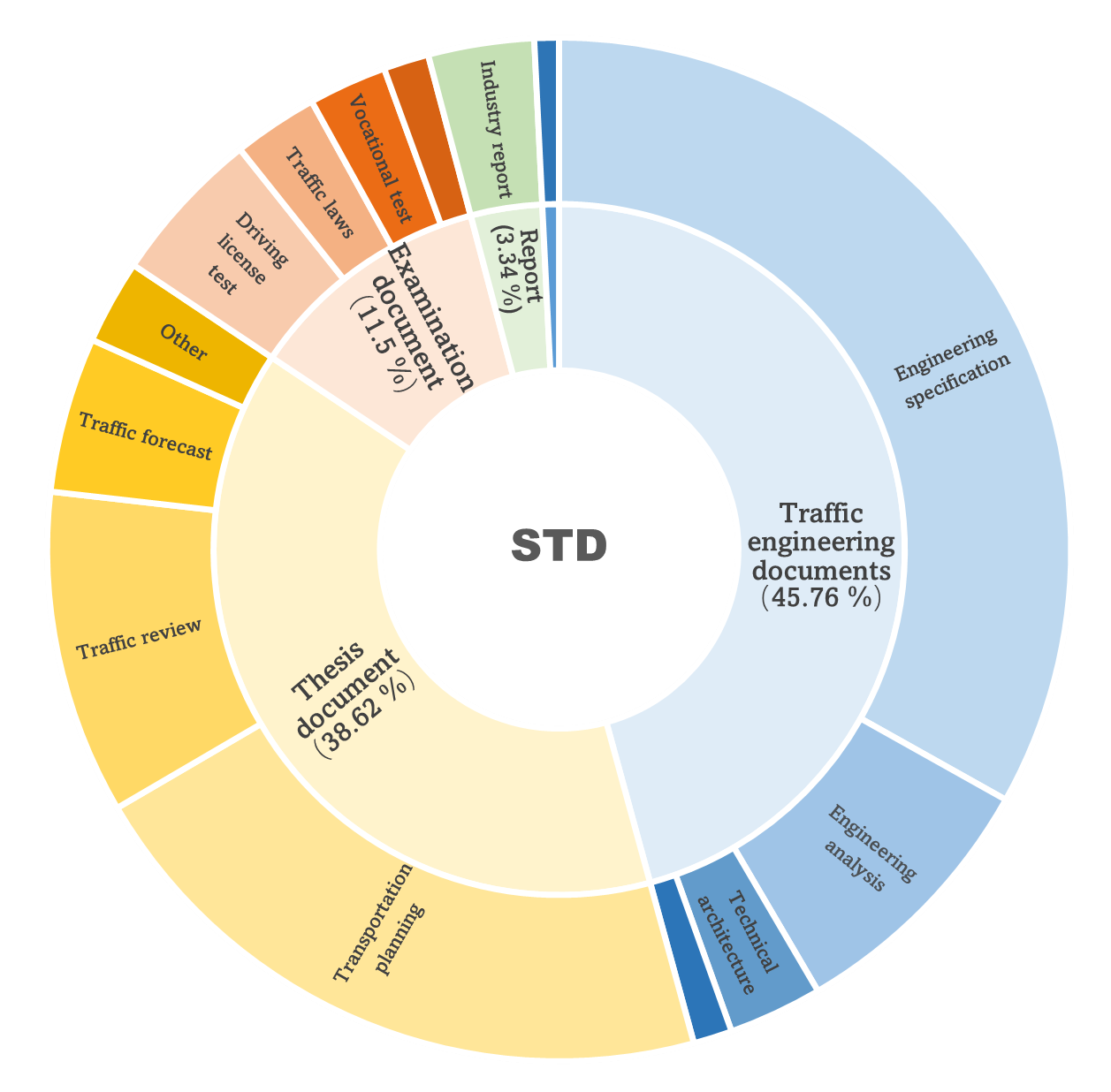}
        \label{fig:data-analysis-std}
    \end{minipage}
    \hfill
    \begin{minipage}{0.48\linewidth}
        \centering
        \includegraphics[width=\linewidth]{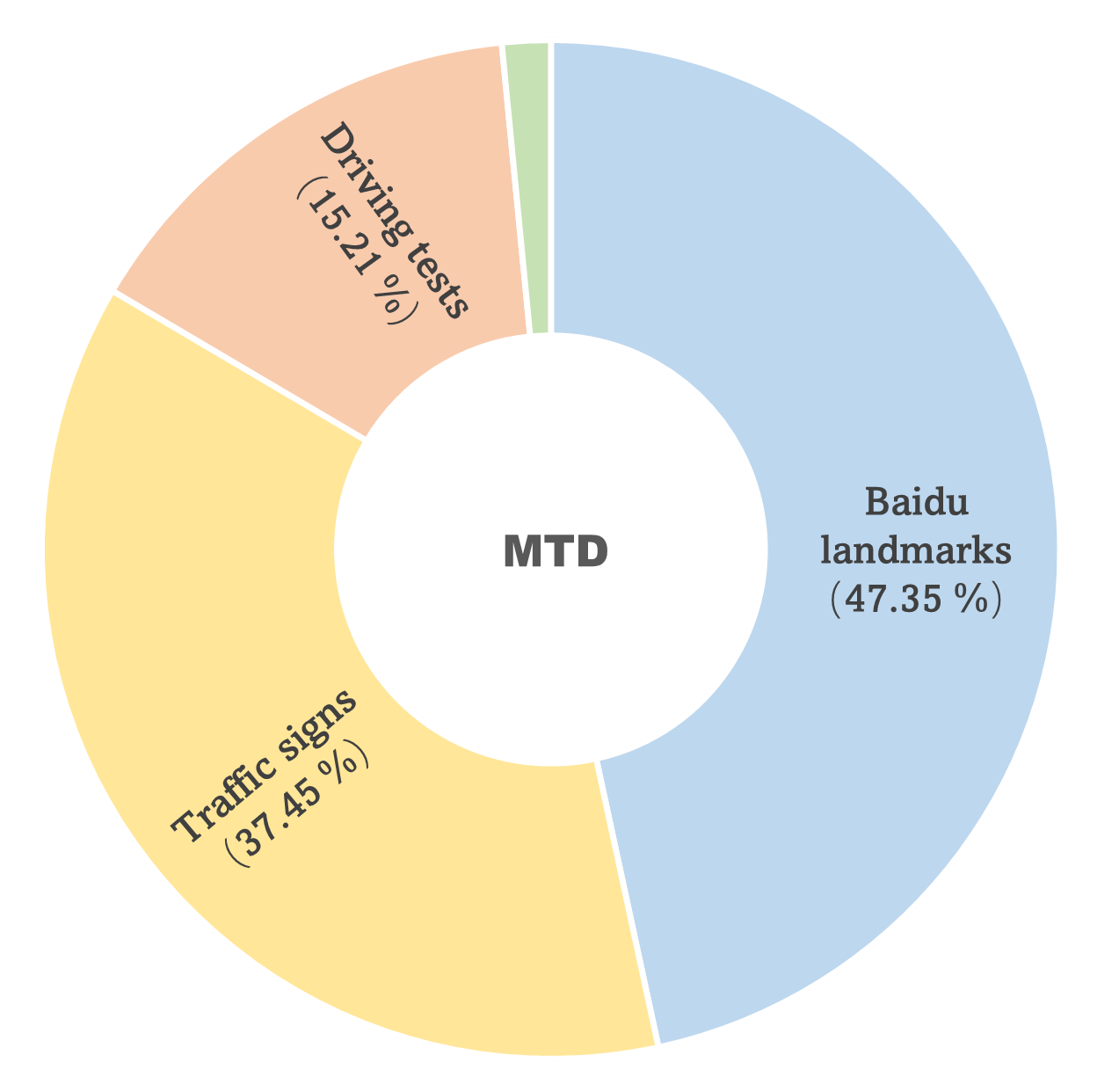}
        \label{fig:data-analysis-mtd}
    \end{minipage}
    \caption{Composition of STD and MTD data.}
    \label{fig:data-analysis}
\end{figure*}

During the data generation process, we organized the question-answer pairs into directories based on the respective categories. Consequently, by calculating and summing the number for each document category, we obtained the composition of data for STD and MTD, as illustrated in Figure \ref{fig:data-analysis}, which presents the statistical information of the dataset we generated. It is evident that STD and MTD encompass various aspects and forms within the transportation domain.

To assess the quality of the generated STD data, we conducted a random sampling method, sampling 200 questions along with their corresponding answers and original text chunks. In order to evaluate the accuracy of each instance, a manual comparison was performed, considering aspects such as the question, answer, and original text. The evaluation results are presented in Table \ref{tab:evaluation-results}, indicating that the majority of samples have clear meaning, although samples may contain some noise within reasonable bounds. Despite the possibility of errors in some generated samples, the majority still maintains the correct format or partial correctness, providing valuable guidance for training models to execute instructions. 

We randomly selected 500 question-answer pairs. Leveraging LLMs, we converted all question-answer pairs into a single-choice format, as illustrated in Figure \ref{fig:data-prompt-example}. After manual reviewing, we identified approximately 300 usable questions. Additionally, we incorporated around 200 traditional multiple-choice questions from the question bank, resulting in a total of about 500 questions. 
\begin{figure*}[htp]
    \centering
    \includegraphics[width=\linewidth]{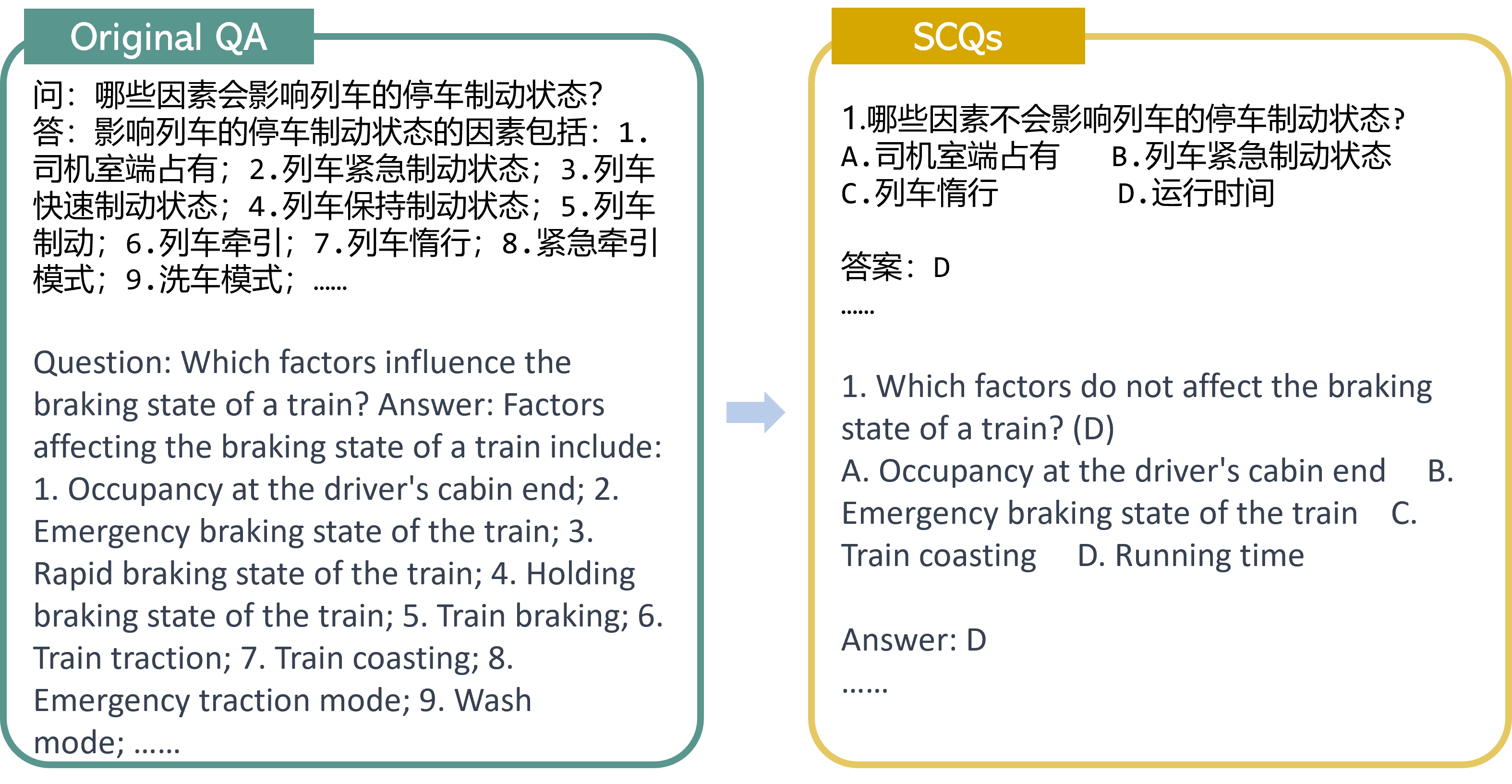}  
    \caption{An example of transforming question-answer pairs into a single-choice format.}
    \label{fig:data-prompt-example}
\end{figure*}

\subsection{Base Model Selection}
In this section, we describe the base model selection process for TransGPT-SM and TransGPT-MM, respectively. We aim to choose the general-purple foundation model that can be finetuned on the domain-specific data and perform well on various tasks in the transportation domain.

There are many existing models that can be used as base models, but not all of them are well-adapted to Chinese as well as English. Therefore, we prioritize the models that have better compatibility with Chinese, such as BLOOMZ~\cite{muennighoff2022crosslingual}, ChatGLM2, Chinese-Alpaca~\cite{cui2023efficient}, etc. These models are large language models for bilingual (Chinese and English) chat.

To evaluate the performance of these models, we use two benchmark datasets: C-Eval~\cite{huang2023c} and CMMLU~\cite{li2023cmmlu}, which are Chinese evaluation datasets for LLMs. These datasets contain multiple-choice questions that test the model's ability to understand and generate natural language in various domains. However, these datasets mainly focus on the selection tasks and do not test the model's ability to generate text and interact with users.
Therefore, we also adopted LLM-benchmark, which is a multi-task evaluation dataset for LLMs in Chinese. This dataset covers a wide range of tasks, such as knowledge question answering, open-ended question answering, numerical calculation, poetry, music, sports, entertainment, writing articles, text translation, code programming, ethical refusal, and multi-turn dialogue. We use Score as the evaluation metric, which is the average score of 100 samples (on a 10-point scale), and use GPT-4 to grade the samples based on the fluency, harmlessness, and correctness of the content. Although the task types are not comprehensive, our evaluation provides a preliminary understanding of the model's performance in various tasks.
The test scores of C-Eval and CMMLU are extracted from the official evaluation or leaderboard submission of each model, while the scores of LLM-benchmark are obtained by running the same evaluation script. The evaluation script prompts follow the official prompts given by each model.
We select five mainstream models and test them on these three datasets. The results are shown in the Table \ref{tab:performance}. 

\begin{table}[ht]
\caption{Performance Comparison of Different Models}
    \centering
    \begin{tabularx}{\textwidth}{CCCC}
        \toprule
        \textbf{Base Model} & \textbf{C-Eval} & \textbf{CMMLU} & \textbf{LLM-Benchmark} \\
        \midrule
        ChatGLM2-6B & 51.7 & 50.0 & 7.7 \\
        Baichuan-7B & 42.8 & 42.3 & 4.5 \\
        BLOOMZ-7B & 37.9 & 37.0 & 3.3 \\
        \footnotesize{Chinese-Alpaca-2-7B} & 42.9 & 41.8 & 8.2 \\
        \footnotesize{Chinese-Alpaca-Plus-7B} & 36.4 & 36.8 & 7.0 \\
        \bottomrule
    \end{tabularx}
    \label{tab:performance}
\end{table}

Based on the comprehensive scores, we choose ChatGLM2-6B as our base model for TransGPT-SM. ChatGLM2-6B, a bilingual (Chinese and English) LLM, has 6 billion parameters.
ChatGLM2-6B has some domain knowledge learned from its pre-training data, which can be useful for our finetuning process.

As for the multi-modal version, we ultimately chose the open-source VisualGLM-6B as the backbone of TransGPT-MM because it has a strong correlation with ChatGLM-6B, which allows for easy subsequent extensions. The language model of VisualGLM-6B is based on ChatGLM-6B and the visual part was aligned to the linguistic modality trough training with BLIP2-Qformer~\cite{li2023blip}.

\subsection{Training}

\paragraph{\textbf{TransGPT-SM}}

To build a LLM tailored for the transportation domain, we employed the technique of Instruction Finetuning (SFT). Utilizing our instruction finetuning dataset, we finetuned the base model. We chose ChatGLM2-6B as the base model, a bilingual LLM supporting both Chinese and English for chat-based scenarios. 
Leveraging this data as guidance, we enhanced the model's ability to adapt to specific knowledge within the transportation domain. We finetuned the model for 3 epochs with a learning rate of 1e-4 and a batch size of 32. We used the Adam optimizer and the cross-entropy loss function. We use the LORA technique following hyper-parameters of \cite{hu2021lora}. 

\paragraph{\textbf{TransGPT-MM}}
We propose a two-stage finetuning approach to enable models to learn domain-specific knowledge while retaining general ability.
In the first stage, we finetune the base model on the combination dataset of our domain-specific MTD and generic CCAC datasets. For each sample, the image and question are used as input and answer as the golden output. We used the LORA~\cite{hu2021lora} technique to finetune the model for 300 epochs with a learning rate of 1e-4, a lora rank of 32 and a batch size of 32. And we used the Adam optimizer and the cross-entropy loss function. In addition, the proportion of specialized and generic samples is roughly the same. So after this phase of finetuning, the model was warmed up, i.e., a small amount of domain-specific knowledge was learned as well as general ability was maintained.
In the second stage, we finetune the model on the domain-specific MTD in order to further gain more knowledge about the transportation domain. In addition to prevent catastrophic forgetting of generic knowledge, 32 generic samples were added. Similar to the first stage, we take the LORA technique and perform 120 epochs training. After this phase of finetuning, we got TransGPT-MM.

\section{Experiments}\label{sec:exp}
We employed the constructed TransEval to investigate whether the unsupervised construction of simulated traffic data (STD) could assist large language models (LLMs) in learning traffic-related knowledge and professional skills. Another aspect to explore is whether injecting domain-specific knowledge would impact the performance in general tasks such as natural language inference (NLI) and question-answering (QA). To address this question, we also evaluated our model on several common benchmarks in both Chinese and English.

We present the results of our single-modal TransGPT-SM and multi-modal TransGPT-MM on the evaluation datasets. 
We compare our models with strong baselines: ChatGLM2-6B and VisualGLM-6B. 

\subsection{Single-modal Experiment}
Considering the close relationship between traffic and engineering projects and the need for rigor in the STD dataset composition, we assessed the model's capabilities in the transportation domain using three tasks for the single-modal model: Traffic Planning Test (TPT), Traffic Engineering Test (TET), and Driving License Exam (DLE).

The Driving License Exam (DLE) serves as a common task to evaluate whether the model comprehends traffic rules and driving skills. It aims to predict correct answers based on driving scenarios and questions. We selected 10 common driving tests and randomly extracted 100 samples for each driving test from the STD.
The Traffic Engineering Test(TET) is a crucial task designed to assess whether the model can understand traffic engineering projects and their associated methods and standards. As traffic planning is a complex task, we focused on testing the model's understanding of knowledge and methods related to traffic planning.

We use TransGPT-SM and baseline model (ChatGLM2-6B) for prediction and evaluation (accuracy) on C-Eval, and TransEval (namely, TPT, TET and DLE). The results in Table \ref{tab:evaluation-results} demonstrate that our model performs significantly better on TET compared to the baseline, with an improvement of 10.16\%. There is a modest enhancement of 1.78\% in DLE, while the difference in TPT is not substantial. In the case of DLE, the limited overlap between our data distribution and DLE results in a less pronounced performance improvement, suggesting a potential need for increased diversity in the dataset. Regarding TPT, due to the inherent depth of knowledge, adopting unsupervised data generation methods may not effectively extract deeper layers of textual information. Considering these three outcomes, it is advisable to explore pre-training knowledge injection approaches and consider expanding data collection through both broadening datasets and manual supplementation to enhance the model's effectiveness.



\begin{table}[htbp]
    \caption{Results of Trans-SM and the bese model.}
    \label{tab:evaluation-results}
    \begin{tabularx}{\textwidth}{CCCCC}
        \toprule
        \textbf{Model} & \textbf{C-Eval}  & \textbf{TET} & \textbf{TPT} & \textbf{DLE} \\
        \midrule
        ChatGLM2-6B & 51.70  & 23.98 & 29.23 & 22.02 \\
        TransGPT-SM & 47.99  & 34.14 & 29.24 & 23.80 \\
        \bottomrule
    \end{tabularx}
\end{table}

\subsection{Multi-modal Experiment}

We evaluate TransGPT-MM and baseline VisualGLM-6B on the testset of MTD by measuring its accuracy in generating natural language outputs based on both textual and visual inputs. We manually judge the predictions, and if the meaning is similar to the golden output, we consider it correct. The results in Table \ref{table:result mm} show that our model outperformed baseline by 40.16\%.

\begin{table}[ht]
\caption{Results of Trans-MM as well as the base model.}
\centering
\begin{tabularx}{\textwidth}{CC}
\toprule
\textbf{Model}	& \textbf{Accuracy}	 \\
\midrule
VisualGLM-6B	            & 27.05\\
TransGPT-MM	            & 67.21\\
\bottomrule
\end{tabularx}
\label{table:result mm}
\end{table}


\section{Analysis}\label{sec:analysis}
\subsection{Ablation Study}
We conduct an ablation study for three components and investigate the impact of different components. We cumulatively remove each component and evaluate it on the test dataset.
-\textit{Filtering} involves the exclusion of the filtering component in Step 5. -\textit{Chunking} involves omitting the introduction of the original text chunk in Step 1. -\textit{Question-Answer} denotes training with the original unlabeled data. We evaluate our model using the TET test dataset. We adopt the single-modal setting.
Results are shown in Table \ref{tab:ablation_study}.
The findings from Table \ref{tab:ablation_study} clearly demonstrate that each component of the data construction process is crucial in enhancing the performance of the zero-shot prompting. Notably, these stages are interdependent and collectively contribute to the overall enhancement achieved. 

\begin{table}[htb!]
    \centering
    \begin{tabular}{lc}
        \toprule
        \multicolumn{1}{c}{\textbf{Method}}                  & \textbf{Accuracy} \\ \hline
        \multicolumn{1}{l|}{All Stages}           & \textcolor{black}{34.14}  \\ 
        \multicolumn{1}{l|}{~~~~~-\textit{Filtering}} & \textcolor{black}{30.48} \\ 
        
        \multicolumn{1}{l|}{~~~~~-\textit{Chunking}}    &     \textcolor{black}{27.23}              \\ 
        
        \multicolumn{1}{l|}{~~~~~-\textit{Question-Answer}}              & \textcolor{black}{23.98}                                                   \\ \bottomrule
    \end{tabular}%
    \caption{Ablation study of components for constructing finetuning data.}
    \label{tab:ablation_study}
\end{table}




\subsection{Case Study}
\begin{figure*}[htp]
    \centering
    \includegraphics[width=2\linewidth/2]{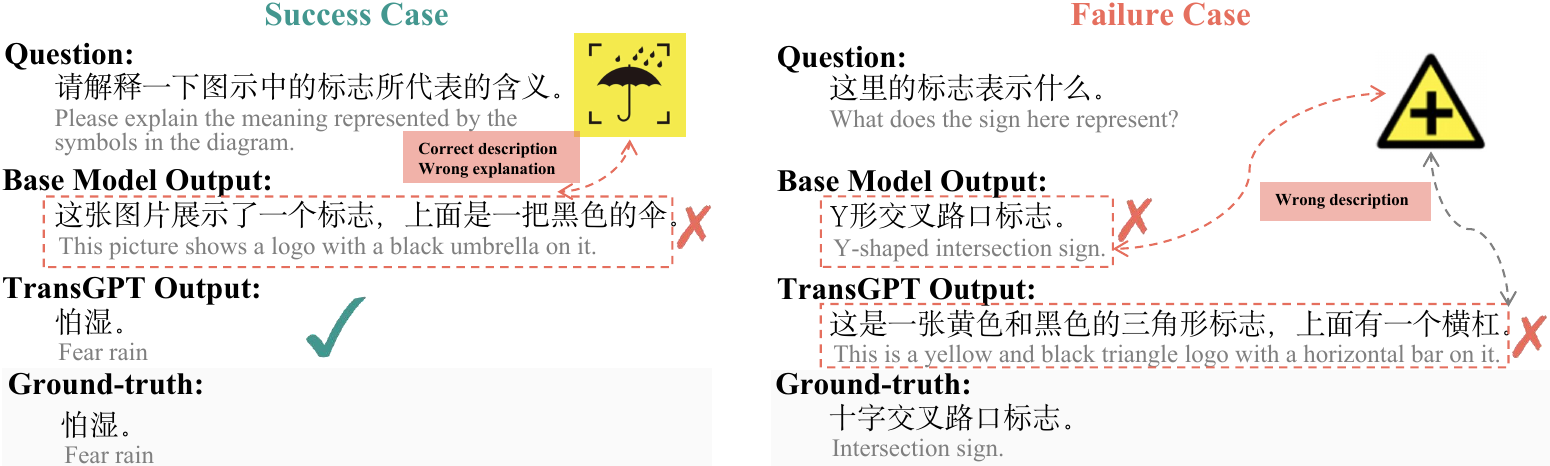}  
    \caption{Case study of TransGPT-MM.}
    \label{fig:case-study}
\end{figure*}
We visualize answers predicted by TransGPT-MM in Figure \ref{fig:case-study}. Success cases are in the left and failure cases are in the right.
For the left case, TransGPT-MM correctly predicts the sign that should be the ``Fear rain''. ``This picture shows a logo with a black umbrella on it.'' is predicted by the base model, which is wrong. The erroneous prediction of the base model is due to the lack of the transportation knowledge. 

For the failed cases, TransGPT-MM and the base model both wrongly predicts the sign. ``Y-shaped intersection sign'' is the ground-truth answer. The erroneous prediction is due to the capability of image understanding. Both models can not correctly recognize the Y shape and + shape.
Interestingly, after combining the question and more knowledge, ``The symbol in the middle of the sign is Y'', as rationales, the predicted result is modified to a correct one. This failure case reveals the importance of rationales, which is future directions to study.

\section{Application}\label{sec:application}
In this section, we will showcase the potential applications of TransGPT for traffic management, such as traffic flow prediction. We use the ASTGCN model as a small module for TransGPT to call as a tool, because the ASTGCN model returns pure text after processing, so we use TransGPT-SM as the main model. We also compare TransGPT-SM+ASTGCN with TransGPT-SM to demonstrate the effectiveness of our model.

\subsection{Generating Synthetic Traffic Scenarios}
One possible application of TransGPT is to predict traffic flow based on some given conditions or constraints. For example, given a location and a time, TransGPT can predict the traffic flow at that location for the next hour. This application can be used for traffic planning and management.

To evaluate this application, we randomly select some locations and times from the Flow forecast dataset, and provide them as inputs to the TransGPT-SM original model and TransGPT-SM+ASTGCN. Then, we compare the generated results with the real data from the Flow forecast dataset. The results are shown in Table 5.

\begin{table}[htbp]
    \caption{Results of Trans-SM and the bese model.}
    \label{tab:evaluation-results}
    \begin{tabularx}{\textwidth}{CCC}
        \toprule
        \textbf{Model} & \textbf{MAE}  & \textbf{RMSE} \\
        \midrule
        TransGPT-SM &  616.36 &  873.27 \\
        TransGPT-SM+ASTGCN & 35.7 & 52.86 \\
        \bottomrule
    \end{tabularx}
\end{table}
As we can see from Table 5, TransGPT-SM+ASTGCN outperforms TransGPT-SM original model in mae and rmse, indicating that TransGPT-SM+ASTGCN can better predict the changes in traffic flow.

\section{Conclusions and Future Direction}\label{sec:conclusion}
TransGPT demonstrates its potential in various applications within traffic analysis and modeling. These applications include generating synthetic traffic scenarios, providing explanations for traffic phenomena, answering traffic-related questions, offering traffic recommendations, and generating comprehensive traffic reports. This groundbreaking work not only contributes to the advancement of natural language processing (NLP) in the transportation field, but also provides a valuable tool for researchers and practitioners in the Intelligent Transportation Systems domain.

The potential of TransGPT extends beyond traffic analysis and modeling, offering valuable insights and applications across various domains related to urban mobility and intelligent transportation systems. For example, TransGPT could be trained on historical traffic data and weather patterns to predict future traffic congestion and suggest preventive measures. This would be particularly useful for major events or unexpected disruptions. We can expect even more innovative ways to leverage this powerful language model for the benefit of individuals, communities, and policymakers.

We believe that there are increasing challenges and opportunities that arise in the context of Transformers within ITS. These challenges may involve obstacles to algorithmic development or exploration (such as benchmark environments or data sources, etc.), while opportunities represent areas of both Transformers or ITS that may require further investigation.
\vspace{6pt} 



\authorcontributions{Conceptualization, P.W.;    methodology, P.W. and X.W.;    software,  P.W.;    validation, P.W. and X.W.;    formal analysis, P.W. and X.W.;    investigation, P.W. and X.W. ;    resources, P.W., X.W., W.H. and F.H.; writing—original draft preparation, P.W., X.W. and W.H. .;  writing—review and editing, P.W., X.W., W.H. and F.H.;  visualization, P.W. and X.W.;
supervision,  P.W. , W.H. and X.W.;  project administration,  P.W.;  funding acquisition,  P.W. All authors have
read and agreed to the published version of the manuscript.}

\funding{This research received no external funding}

\begin{adjustwidth}{-\extralength}{0cm}

\reftitle{References}

\bibliography{reference}

\end{adjustwidth}
\end{document}